\documentclass{bmvc2k}

\usepackage{amsmath,amssymb,amsthm}
\usepackage{booktabs}
\usepackage{multirow}
\usepackage{xcolor}
\usepackage{tikz}
\usepackage{capt-of}
\usepackage{placeins}
\usepackage[capitalize,noabbrev]{cleveref}
\usepackage{algorithm}
\usepackage{algpseudocode}
\hypersetup{hypertexnames=false}

\title{Off-Manifold Refinement: Guiding Video Generators with a Frozen World Model}

\addauthor{Hai Nguyen-Truong}{nguyentruonghai141@gmail.com}{1}
\addauthor{Tuan-Anh Vu}{tuananh.vu@ucla.edu}{2}
\addauthor{Dang Huynh}{dang.huynh@fulbright.edu.vn}{1}

\addinstitution{
 Fulbright University Vietnam
}
\addinstitution{
 University of California, Los Angeles
}

\runninghead{Nguyen-Truong et al.}{Off-Manifold Refinement}

\newcommand{\methodname}{OMR}
\newcommand{\methodfull}{Off-Manifold Refinement}
\newcommand{\vjepa}{V-JEPA~2.1}

\newcommand{\zt}{z_{t}}
\newcommand{\zhone}{\hat{z}_{1}}
\newcommand{\uth}{u_{\theta}}
\newcommand{\Evj}{E_{\mathrm{vjepa}}}
\newcommand{\Aps}{A_{\psi}}
\newcommand{\Eph}{E_{\phi}}
\newcommand{\Pph}{P_{\phi}}
\newcommand{\algbase}[1]{\textcolor{black!65}{#1}}
\newcommand{\algomr}[1]{\textcolor{blue!65!black}{\textbf{#1}}}
\newcommand{\algomrmath}[1]{\textcolor{blue!65!black}{#1}}

\begin{document}
\maketitle

\begin{abstract}
Modern video generators routinely fail at physical dynamics --- objects float, trajectories violate gravity, contacts vanish. Standard denoising and flow-matching objectives fit visual data distributions but do not explicitly penalize such physical violations. Existing remedies can improve physical consistency, but typically add substantial inference or training cost. Candidate-selection methods generate and score multiple videos, while gradient-based world-model guidance repeatedly decodes and re-encodes intermediate estimates. Generator-internal refinement adds perturbation and re-denoising loops, whereas post-training requires curated data and additional optimization. We propose \textbf{\methodfull{} (\methodname{})}, an inference-time method that instead injects world-model feedback directly into a single sampling trajectory. During scheduled middle ODE steps, we augment the generator velocity with the gradient of an adapter-space \vjepa{} surprise energy. This external correction can move the latent away from the uncorrected sampling trajectory and toward regions ranked as more physically plausible by the frozen predictor, after which the generator continues rendering from the corrected state. A small trained latent-to-embedding adapter keeps the gradient tractable at inference, and both the video generator and the world model remain frozen. On our fixed $400$-prompt VideoPhy-2 detailed subset, \methodname{} lifts the joint Semantic-Adherence-and-Physical-Commonsense metric from $47.0\%$ to $52.0\%$ ($+5.0$pp absolute, $+10.6\%$ relative) over the base Wan2.2-T2V-A14B sampler. On a separate fixed $50$-prompt efficiency subset, it requires $1.71\times$ the base runtime rather than the multiplicative cost of reward/search alternatives. \textbf{Project Page:} \url{https://itruonghai.github.io/omr}.
\end{abstract}

\vspace{-2mm}
\section{Introduction}
\vspace{-2mm}
\label{sec:intro}

State-of-the-art video generators such as Wan2.2~\citep{wan22} and world-foundation-model platforms such as Cosmos~\citep{cosmos} produce videos of remarkable visual quality. Yet recent benchmarks and physics-alignment studies show that generated videos still routinely violate basic physical laws~\citep{videophy2,yuan2026wmreward,le2025newtonrewards}: objects float against gravity, fluid surfaces lose continuity, hands pass through objects they should grasp, and collisions ignore material properties such as deformation, cracking, or rebound. Standard flow-matching~\citep{lipman2023flowmatching} and denoising-diffusion objectives reconstruct the visual statistics of training videos but do not explicitly penalize these physical violations. Consequently, a generator can learn \emph{what the world looks like} without reliably learning \emph{how the world behaves}.

\begin{figure}[t]
    \centering
    \includegraphics[width=\linewidth]{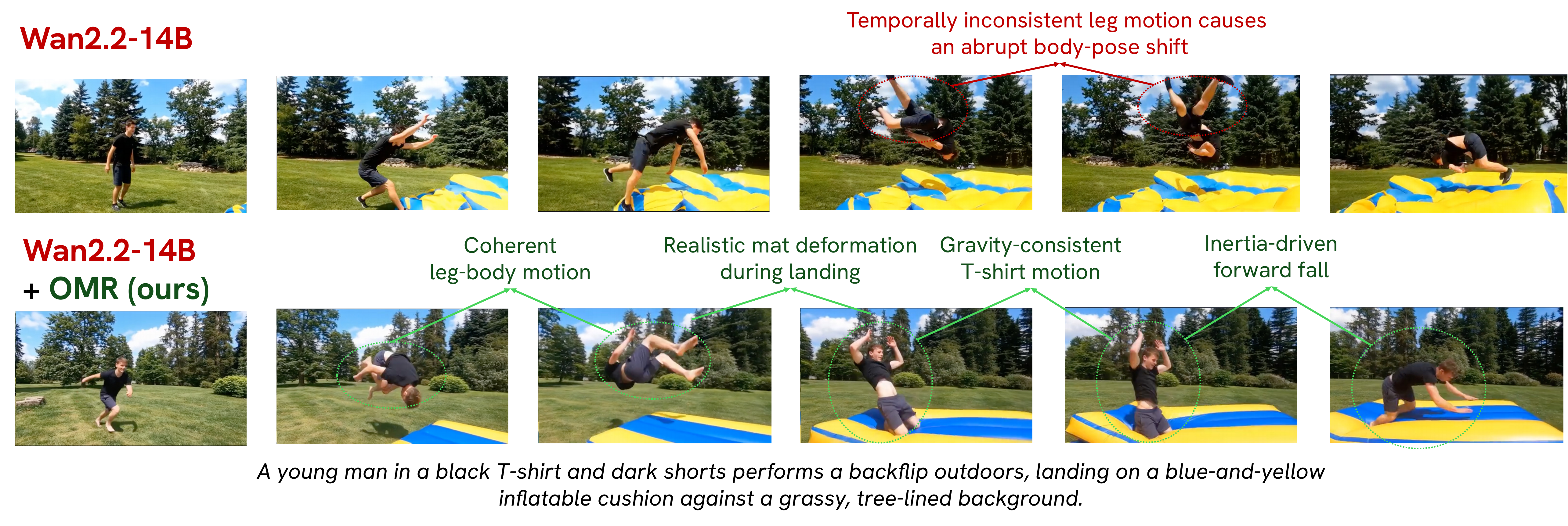}
    \vspace{-8mm}
    \caption{\textbf{\methodname{} improves physical plausibility.}
    Given the same Wan2.2 generator, the baseline produces temporally inconsistent leg motion and an abrupt body-pose shift during a backflip landing, while \methodname{} yields coherent leg-body motion, mat deformation, gravity-consistent clothing motion, and inertia-driven fall.}
    \label{fig:beforeteaser}
    \vspace{-4mm}
\end{figure}

\Cref{fig:beforeteaser} makes this gap concrete. The baseline frames are individually plausible, but the landing sequence contains the kind of temporal physics error that is easy to miss from still-image quality alone: body pose changes abruptly, limb motion is inconsistent, and the cushion does not respond coherently to impact. With \methodname{}, the same generator and prompt produce a more physically coherent continuation, with leg-body motion, material deformation, clothing motion, and inertia moving in the same direction. This is the capability we target most directly: improving local physical interactions inside the generated trajectory rather than merely polishing appearance.

Recent attempts to improve physical consistency intervene either during training or inference. Post-training methods modify the generator through physics-aware conditioning, distillation, or reward optimization~\citep{wang2025wisa,zhang2025videorepa,wang2025physcorr,le2025newtonrewards,an2026vggrpo,kupyn2025epipolar}, but require additional data and compute and bind the improvement to a particular checkpoint. Inference-time methods keep the generator frozen. Candidate-selection approaches score multiple completed videos~\citep{yuan2025vjepa2reward,yuan2026wmreward}, so their cost grows with the number of generated candidates; in-trajectory approaches instead steer an unfinished sample, but existing methods rely on repeated perturbation and re-denoising~\citep{jang2026selfrefining} or repeatedly decode intermediate clean estimates to pixels and re-encode them for world-model evaluation~\citep{yuan2026wmreward}. These limitations motivate an efficient way to apply world-model guidance directly within a single sampling trajectory.

\begin{figure}[t]
    \centering
    \includegraphics[width=\linewidth]{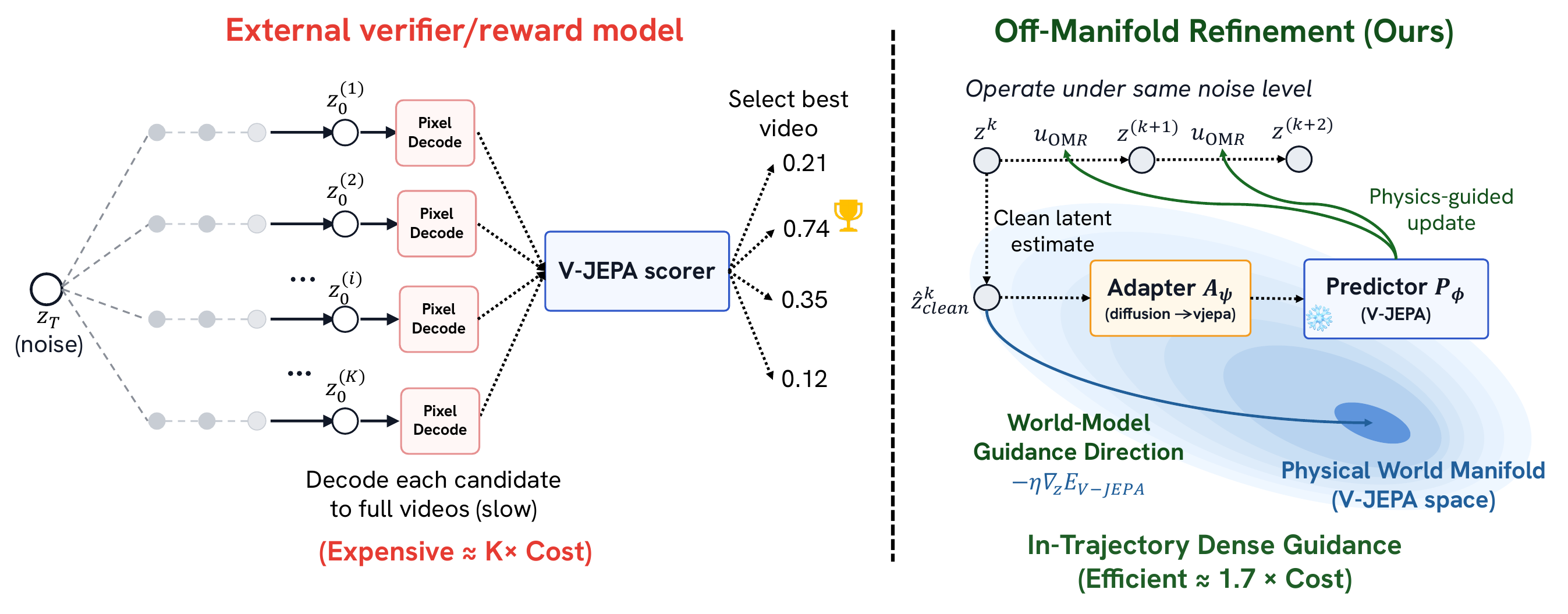}
    \vspace{-8mm}
    \caption{\textbf{\methodname{} spends the world-model signal inside the sampling trajectory rather than after generation.} Left: the candidate-selection variants of verifier or reward methods~\citep{yuan2025vjepa2reward,yuan2026wmreward} decode $K$ completed candidate videos and use a \vjepa{}~\citep{vjepa21} scorer only to choose among finished samples, so cost scales with the number of candidates. Right: \methodname{} stays within a single ODE trajectory: an adapter maps the clean latent estimate into \vjepa{} space, the frozen predictor defines a dense surprise energy, and its gradient supplies a physics-guided update before the final video is decoded.}
    \vspace{-5mm}
    \label{fig:teaser}
\end{figure}

\Cref{fig:teaser} summarizes the paper's central contrast. In the candidate-selection branch, a verifier pipeline asks a world model to judge completed candidate videos, so every additional use of the physical signal costs another full generation. \methodname{} asks a different question: can we spend that same world-model signal \emph{before} the sample is finished, while the generator trajectory is still movable? A standard ODE step follows the generator's learned vector field; \methodname{} adds an adapter-space \vjepa{}~\citep{vjepa21} surprise force whose gradient need not align with that field. This is the off-manifold refinement: the correction can move the latent away from a physically implausible uncorrected trajectory, after which subsequent generator steps continue rendering from the corrected state.

We propose \methodfull{} (\methodname{}), an inference-time procedure that uses a frozen video world model as a dense differentiable guidance signal inside one sampling trajectory rather than as a scalar score over many completed clips. At each ODE sampling step, we compute the generator's velocity as usual; during the scheduled middle steps, we pass the implicit clean prediction $\hat{z}_1$ through a small trained adapter $A_\psi$ that maps generator latents directly into \vjepa{} embeddings~\citep{vjepa21}. This adapter \emph{replaces \vjepa{}'s encoder at sampling time} --- only the predictor $P_\phi$ runs --- so an adapter-space surprise gradient can be computed without ever decoding to pixels or running the encoder. We subtract that gradient only at the scheduled steps; the remaining steps use the frozen generator velocity alone. Because this guidance path avoids full candidate generations and bypasses both pixel decoding and the \vjepa{} encoder, \cref{tab:wallclock} measures a relative runtime of $1.71\times$ on a fixed $50$-prompt efficiency subset, not the multiplicative cost of best-of-$K$ or rejection sampling.

\vspace{-2mm}
\paragraph{Why ``off-manifold''?}
We use ``manifold'' as operational shorthand for the local sampling geometry preferred by the frozen generator's ODE field, not as a claim that one velocity vector spans a formal data-manifold tangent space. A generator-only refinement step follows this learned geometry by asking the generator to denoise or self-consistently resample again. \methodname{} instead differentiates an adapter-space predictive loss from a separately trained world model. Because that gradient is not a likelihood score of the generator, it need not align with the uncorrected generator trajectory. The update is therefore off-manifold in this operational sense, while later generator velocities continue rendering from the corrected latent.

\label{sec:what-not}
Three nearby approaches deserve preemption. (i)~\emph{\methodname{} is not classifier guidance.} The structural form $u_{\text{total}} = u_\theta - \eta \nabla E$ mirrors classifier guidance~\citep{dhariwal2021guidance}, but \methodname{} requires no conditioning label, uses a dense per-patch self-supervised signal rather than a single discriminative scalar, and trains only a small embedding adapter rather than a classifier head on noisy diffusion representations. (ii)~\emph{\methodname{} is not WMReward-style trajectory steering.} Recent work uses VJEPA-2 as a scalar reward for best-of-$K$, rejection, and trajectory steering~\citep{yuan2025vjepa2reward,yuan2026wmreward}; \methodname{} instead differentiates an adapter-space predictive objective inside a single trajectory, avoiding both in-loop VAE decoding and the \vjepa{} encoder. Our direct comparison is therefore near-budget best-of-$K$ and reward-guidance baselines, not only the base generator. (iii)~\emph{\methodname{} is not self-refining video sampling.} Self-Refining Video Sampling~\citep{jang2026selfrefining} obtains a zeroth-order uncertainty signal by perturbing and re-denoising within an outer ODE step, then using cross-sample disagreement to decide which regions to refine. \methodname{} has no perturbation step, no re-noising, and no mask blend: the adapter-space predictive gradient supplies the correction direction in a single monotone trajectory.
\paragraph{Contributions.} We summarize the main contributions and findings.
\begin{itemize}
    \item \textbf{\methodfull{} (\methodname{})}: an inference-time sampler that keeps the generator and world model frozen while adding one adapter-space surprise-gradient correction to each scheduled middle ODE step.
    \item \textbf{Efficient encoder-free guidance}: a $33.3$M convolutional latent-to-embedding adapter maps generator latents directly into world-model embedding space, avoiding in-loop VAE decoding and video-encoder passes while keeping the sampler close to single-trajectory cost.
    \item \textbf{Evidence across two generators and two benchmarks}: on our fixed $400$-prompt VideoPhy-2 detailed subset, \methodname{} raises Wan2.2 joint score from $47.0\%$ to $52.0\%$; a separate fixed $50$-prompt efficiency study measures $1.71\times$ relative runtime. In fixed $50$-prompt pilots, it raises CogVideoX-5B VideoPhy-2 from $32.0\%$ to $42.0\%$ and Wan2.2-I2V Physics-IQ from $26.44$ to $31.34$.
    \item \textbf{Broad physical-consistency gains}: \methodname{} yields non-decreasing PC and joint scores in every reported category. Joint gains are largest in contact dynamics ($+11.0$pp) and fluid/deformable motion ($+9.8$pp).
\end{itemize}

\vspace{-4mm}
\section{Related Work}
\vspace{-2mm}
\label{sec:related}

\paragraph{Diffusion and flow-based video generation.}
Modern text-to-video systems build on diffusion and score-modeling methods such as DDPMs, score SDEs, DDIM, and improved diffusion training~\citep{ho2020ddpm,song2021scorebased,song2021ddim,nichol2021improved,dhariwal2021guidance}. High-resolution generation has benefited from latent diffusion, improved noise schedules, transformer backbones, consistency distillation, and flow matching~\citep{rombach2022ldm,karras2022edm,peebles2023dit,song2023consistency,lipman2023flowmatching}; video models add temporal structure through video diffusion objectives, temporal modules, or image-to-video adaptation~\citep{ho2022video,ho2022imagenvideo,singer2022makeavideo,blattmann2023videoldm,blattmann2023svd,guo2024animatediff}. These advances improve visual quality, but their likelihood-like objectives do not directly enforce physical consistency. \methodname{} targets this remaining gap at inference time.
OneStory~\citep{an2025onestory} instead addresses coherent multi-shot generation by fine-tuning an image-to-video model on a curated dataset and carrying selected visual memories across shots. Its narrative-continuity objective is complementary to the within-trajectory physical guidance studied here.

\paragraph{Physics-aware video generation.}
Recent work injects physical signal either by changing the generator or by applying an external signal at inference time. WISA adds explicit physical descriptions and categorical controls~\citep{wang2025wisa}, VideoREPA distills video-foundation-model relations into a T2V model~\citep{zhang2025videorepa}, and preference/reward approaches train physics-aware reward or DPO objectives~\citep{wang2025physcorr,le2025newtonrewards}. VGGRPO~\citep{an2026vggrpo} post-trains a generator with a latent 4D-geometric reward, while Epipolar Geometry~\citep{kupyn2025epipolar} uses preference-based post-training to improve camera and geometric consistency through pairwise epipolar constraints. Both modify the generator; their geometry-focused training signals are complementary to, but operationally distinct from, \methodname{}'s frozen-generator inference-time correction. In parallel, VJEPA-style world models have been used as verifiers or rewards for best-of-$K$, rejection, and trajectory steering~\citep{yuan2025vjepa2reward,yuan2026wmreward}. These methods show that external physical priors are useful, but they either require generator adaptation or incur a decode-and-encode path when scoring completed candidates or steering an intermediate clean estimate. \methodname{} keeps the generator frozen and spends the world-model signal inside a single trajectory through an encoder-free adapter path.

\paragraph{Inference-time guidance.}
Classifier and classifier-free guidance modify sampling with external or conditional gradients~\citep{dhariwal2021guidance,ho2022classifierfree}; later guidance methods generalize this idea to arbitrary differentiable losses, inverse-problem likelihoods, restarts, or reward functions~\citep{bansal2023universal,chung2023dps,yu2023freedom,restart}. \methodname{} follows this broad template but uses a self-supervised video world model rather than a classifier or learned scalar reward. A complementary no-generator-fine-tuning competitor is Self-Refining Video Sampling~\citep{jang2026selfrefining}, which obtains a zeroth-order signal by repeatedly perturbing and re-denoising uncertain regions. \methodname{} instead uses one adapter-space predictive-surprise gradient at each scheduled guided step of a monotone trajectory.

\paragraph{Video world models.}
World models learn compact predictive dynamics for planning and control~\citep{ha2018worldmodels,hafner2020dreamer,lecun2022path}. JEPA-family video models~\citep{vjepa,vjepa2,vjepa21} bring this predictive idea to large-scale self-supervised representation learning over masked space-time patches. We use \vjepa{} because it provides dense patch features and a frozen predictor head, which together define the adapter-space surprise energy used by \methodname{}.



\vspace{-4mm}
\section{Method}
\vspace{-2mm}
\label{sec:method}

\subsection{Background}
\label{sec:background}

\paragraph{Flow-matching video generation.}
Throughout the paper, $t$ denotes \emph{clean time}: $t=0$ is pure noise and $t=1$ is a clean video latent. A flow-matching video generator~\citep{lipman2023flowmatching} learns a velocity field $\uth(\zt,t)$ that transports latents along this noise-to-video path:
\begin{equation}
    \frac{d\zt}{dt} = \uth(\zt, t), \quad t \in [0, 1]
    \label{eq:flow-ode}
\end{equation}
The same velocity field also estimates the final clean latent. From any intermediate state $\zt$, the model predicts where the latent would land at $t=1$ under a constant-velocity extrapolation:
\begin{equation}
    \zhone^{(\theta)}(\zt, t) = \zt + (1-t) \cdot \uth(\zt, t).
    \label{eq:clean-pred}
\end{equation}
This estimated latent is the object on which \methodname{} evaluates the world model correction.
\vspace{-2mm}
\paragraph{\vjepa{} as a video world model.}
\vjepa{}~\citep{vjepa21} has an encoder $\Eph$ and a predictor $\Pph$ that forecasts masked future embeddings from visible context. We score a $48$-frame window from each $81$-frame video. Let $e=\Eph(v)$, with $C$ the visible context tokens, $M$ the masked future targets, and $\delta_M$ learned mask tokens carrying their spatiotemporal positions. Predictive surprise is
\begin{equation}
    \Evj(v) = \big\| \Pph(e_C,\delta_M) - e_M \big\|_2^2 .
    \label{eq:vjepa-energy}
\end{equation}
Low energy means that the masked future is predictable from context in a representation trained on real videos. The energy is dense, differentiable, and depends on embeddings rather than how they were produced.

\subsection{The \methodname{} update rule}
\label{sec:update}

Standard ODE sampling integrates the generator's learned velocity:
\begin{equation}
    z_{t+\Delta t} \leftarrow \zt + \Delta t \cdot \uth(\zt, t).
    \label{eq:std-update}
\end{equation}

\methodname{} augments this with a world-model correction at the scheduled steps for which $\eta(t)>0$. We first compute the implicit clean prediction $\zhone$ via \cref{eq:clean-pred}. We then map it directly into \vjepa{}'s~\citep{vjepa21} embedding space using the latent-to-embedding adapter $\Aps$ (\cref{sec:adapter}):
\begin{equation}
    e \;=\; \Aps(\zhone) \;\in\; \mathbb{R}^{N \times D}, \quad N=T_{\mathrm{tok}}H_pW_p .
    \label{eq:adapter-out}
\end{equation}
At inference time the adapter \emph{replaces} \vjepa{}'s~\citep{vjepa21} encoder $\Eph$ entirely; only the predictor $\Pph$ runs. This is the central efficiency move of the method --- it eliminates the VAE decode and the encoder forward from both the forward and backward passes, with the measured wall-clock effect reported in \cref{tab:wallclock}. The inference-time predictive energy is
\begin{equation}
    \Evj^{\mathrm{OMR}}(\zhone) \;=\; \big\| \Pph\big(e_C,\delta_M\big) \;-\; e_M \big\|_2^2,
    \label{eq:omr-energy}
\end{equation}
where we reshape the adapter tokens to the \vjepa{} grid and apply the same $48$-frame window and mask as the frozen predictor. The gradient with respect to the clean prediction is
\begin{equation}
    g(\zt, t) \;=\; \nabla_{\zhone}\, \Evj^{\mathrm{OMR}}(\zhone),
    \label{eq:omr-grad}
\end{equation}
computed by a single autograd pass through $\Pph$ and $\Aps$. We treat $\zhone$ as the guidance variable and do not backpropagate through the base generator's velocity network $\uth$; equivalently, $\uth$ supplies the ODE velocity and $\nabla_{\zhone}\Evj^{\mathrm{OMR}}$ supplies a local latent correction in the same coordinates. Because the encoder $\Eph$ is not in the path, this backward pass is dominated by the (relatively small) predictor and the (small) adapter; the heavy components --- the VAE decoder and the \vjepa{}~\citep{vjepa21} encoder --- never appear.

The \methodname{} update becomes
\begin{equation}
    z_{t+\Delta t} \;\leftarrow\; \zt \;+\; \Delta t \cdot \uth(\zt, t) \;-\; \eta(t) \cdot g(\zt, t),
    \label{eq:omr-update}
\end{equation}
with $\eta(t)$ a time-dependent guidance strength (\cref{sec:schedule}). We write the correction as an operator-splitting step after the Euler velocity update, so the step-size dependence of guidance is absorbed into the calibrated scalar $\eta(t)$; equivalently, an implementation may parameterize $\eta(t)=\Delta t\,\lambda(t)$. Many Wan-style implementations expose the reverse noise level $\sigma=1-t$, where $\sigma=1$ is noise and $\sigma=0$ is clean. Our equations stay in clean time $t$; an implementation using $\sigma$ should substitute $t=1-\sigma$ and convert step sizes accordingly.

\subsection{Mechanistic role of the correction}
\label{sec:omr-mechanics}

\vjepa{}'s~\citep{vjepa21} predictor is trained on real videos to forecast masked future token embeddings from visible context tokens and their target positions. This objective gives a learned prior over plausible temporal continuations: objects should persist, motion should remain smooth, and contacts should produce consistent downstream effects. When the generator proposes a future segment that does not follow from its context under this prior, $\Pph(e_C,\delta_M)$ diverges from $e_M$ and the energy in \cref{eq:omr-energy} rises. The gradient $g$ is therefore not a generic visual-quality gradient; it is the local direction, in generator-latent coordinates, that most reduces the world model's predictive disagreement for the current clean-latent estimate.

This is the sense in which the update is \emph{off-manifold}. A pure flow-matching step follows the generator's learned vector field $\uth$, which defines the local uncorrected trajectory. The adapter-space surprise gradient comes from a separately trained world model and need not align with that instantaneous flow direction. Subtracting $\eta(t)g$ can move the latent away from a physically implausible uncorrected state; subsequent ODE steps then use $\uth$ to continue rendering from the correction. The method therefore does not replace the generator with \vjepa{}; it adds a small external physical prior to the generator's own dynamics.

\subsection{The latent-to-embedding adapter}
\label{sec:adapter}

Computing $g(\zt, t)$ naively requires decoding $\zhone$ to pixels via the generator's VAE, encoding through \vjepa{}'s encoder, computing the energy, and backpropagating through both. This is prohibitive: even a single backward pass through the VAE on a 16-frame $480 \times 832$ clip exceeds the memory of a single 96\,GB GPU. The adapter $\Aps : Z_{\mathrm{lat}} \to Z_{\mathrm{vjepa}}$ short-circuits both the decode and the encoder by learning, once and offline, to map generator latents directly to the \vjepa{}~\citep{vjepa21} embedding space that the encoder would have produced from the decoded pixels.

\paragraph{Instantiation and training.} For Wan2.2~\citep{wan22}, $\Aps$ is a $33.3$M convolutional adapter trained offline on a $\sim64{,}000$-clip OpenVid-1M~\citep{nan2025openvid} subset. We cache each clip's Wan2.2 VAE latent and frozen \vjepa{}~\citep{vjepa21} encoder target, then train on the paired tensors. For CogVideoX, we retrain the same $33.3$M architecture on CogVideoX VAE-latent/\vjepa{} pairs from the same clips; weights are not shared. Architecture and tensor details are supplementary (\cref{app:adapter}).

\paragraph{Real-video training, generated-video use.} The primary adapter is trained on latents obtained by encoding real OpenVid-1M clips, but \methodname{} uses it during generation on the provisional clean prediction $\zhone$ inferred from a noisy synthetic sampling state $\zt$. This is a genuine distribution shift, but it is also the intended use case: $\Aps$ is not asked to classify whether a latent is real or generated; it is asked to express any Wan2.2 latent in the coordinate system of the frozen \vjepa{} encoder. We reduce the hardest part of the shift in two ways. First, the adapter is applied to $\zhone$, the generator's current estimate of the clean video latent, rather than to raw high-noise $\zt$. Second, the guidance schedule in \cref{sec:schedule} avoids very early steps where $\zhone$ is mostly noise and the adapter output is least reliable. The remaining question is empirical, so \cref{tab:adapter-fidelity} checks representation and surprise-ranking fidelity, and \cref{tab:videophy2,tab:wallclock} show that the frozen Wan2.2 adapter gives downstream gains on generated samples.

\paragraph{Adapter objective.} The adapter is not trained to produce pixels or to judge video quality. It is trained to reproduce the representation that the frozen \vjepa{} encoder would have produced after decoding the same Wan2.2 latent. Let $e\in\mathbb{R}^{N\times D}$ be the frozen encoder target and $\hat{e}=\Aps(z)$ be the adapter prediction, with $N$ space-time tokens of dimension $D$. We normalize each token as $\bar e_i=e_i/(\|e_i\|_2+\epsilon)$ and $\bar{\hat e}_i=\hat e_i/(\|\hat e_i\|_2+\epsilon)$. The mean token cosine is
\begin{equation}
    \cos(\hat e,e)
    =
    \frac{1}{N}\sum_{i=1}^{N}\bar{\hat e}_i^\top \bar e_i ,
    \label{eq:adapter-cos}
\end{equation}
and the token-token relation matrix is
\begin{equation}
    \mathrm{Sim}(e)_{ij}=\bar e_i^\top \bar e_j ,
    \qquad
    \mathrm{Sim}(\hat e)_{ij}=\bar{\hat e}_i^\top \bar{\hat e}_j .
    \label{eq:adapter-sim}
\end{equation}
The adapter loss combines mean squared reconstruction, mean per-token directional alignment, and mean absolute relational-geometry matching:
\begin{equation}
    \mathcal{L}_{\mathrm{adapter}}
    =
    \frac{1}{ND}\|\hat{e} - e\|_2^2
    + 0.5\,(1-\cos(\hat{e},e))
    + \frac{0.1}{N^2}\,\|\mathrm{Sim}(\hat{e})-\mathrm{Sim}(e)\|_1 .
    \label{eq:adapter-loss}
\end{equation}
The first two terms match the coordinates and directions of individual \vjepa{} tokens. The $\mathrm{Sim}$ term is a label-free relational distillation term: it preserves pairwise token neighborhoods so the frozen predictor sees a coherent context--future geometry rather than independently fitted tokens. Its coefficient is small because it is a stabilizer, not a separate physics reward. A longer explanation is provided in the supplementary material (\cref{app:adapter-loss}); the main text uses the adapter as the mechanism that makes adapter-space guidance tractable.

\paragraph{Why the adapter is load-bearing.} Because \methodname{} follows an adapter-space signal at inference time, $\Aps$ determines what information actually reaches the generator latent. The validation should therefore check both representation fidelity and guidance-order fidelity: the predicted tokens should remain aligned with the reference decode-and-encode \vjepa{}~\citep{vjepa21} embeddings, and adapter-space surprise should preserve the plausibility ordering induced by the reference path. We report both checks in \cref{sec:adapter-fidelity}: on the same $81$-frame $480{\times}832$ setting used in the main experiments, the adapter reaches strong token-level cosine agreement and positive rank correlation with reference \vjepa{} surprise. We therefore treat $\Aps$ as the mechanism that makes dense world-model guidance practical during sampling.

\paragraph{Frozen at inference.} The generator-specific adapter, generator, and \vjepa{}~\citep{vjepa21} remain frozen at inference. We train only the adapter, once per generator/VAE latent space.

\subsection{The guidance schedule}
\label{sec:schedule}

The guidance strength $\eta(t)$ is not constant. Under the clean-time convention above, very small $t$ means high noise and very large $t$ means nearly clean. At high noise, the implicit clean prediction $\zhone$ is unreliable; gradient signals computed on it are dominated by noise and VAE artifacts rather than physics. Near $t=1$, most spatial and temporal structure is already locked in, and additional correction has diminishing returns. We concentrate guidance in a middle clean-time band:
\begin{equation}
    \eta(t) \;=\; \eta_0 \cdot \mathbf{1}\big[t_\mathrm{lo} \leq t \leq t_\mathrm{hi}\big],
    \label{eq:schedule}
\end{equation}
with $\eta_0$, $t_\mathrm{lo}$, and $t_\mathrm{hi}$ chosen by a coarse sweep on a held-out calibration set and then fixed for all reported test prompts. Unlike substep-based methods, the schedule is defined directly over the outer ODE timesteps; there is no interaction with an inner refinement loop because no inner loop exists.

\subsection{Algorithm}
\label{sec:algo}

\begin{algorithm}[t]
\caption{\methodname{} sampling}
\label{alg:omr}
\begin{algorithmic}[1]
\Require frozen generator $\uth$, frozen predictor $\Pph$, frozen adapter $\Aps$, clean-time steps $0=t_0<\cdots<t_N=1$, guidance $\eta(\cdot)$, visible/target indices $C,M$, mask tokens $\delta_M$
\State sample $z_{t_0} \sim \mathcal{N}(0, I)$
\For{$i = 0, \ldots, N-1$}
    \State $\Delta t \gets t_{i+1} - t_i$
    \Statex \algbase{\emph{Standard flow-matching prediction}}
    \State $v \gets \uth(z_{t_i}, t_i)$ \Comment{generator velocity}
    \State $\zhone \gets z_{t_i} + (1 - t_i)\, v$ \Comment{clean-latent estimate; stop grad through $\uth$}
    \If{$\eta(t_i) > 0$}
        \State $e \gets \Aps(\zhone)$ \Comment{\algomr{OMR:} adapter replaces VAE decode + \vjepa{} encoder}
        \State $E \gets \big\| \Pph(e_C,\delta_M) - e_M \big\|_2^2$ \Comment{masked-future surprise}
        \State $g \gets \nabla_{\zhone}\, E$ \Comment{one backward pass through $\Pph$ and $\Aps$}
    \Else
        \State $g \gets 0$
    \EndIf
    \Statex \algomr{Guided Euler step}
    \State $z_{t_{i+1}} \gets z_{t_i} \;+\; \Delta t \cdot v \;\algomrmath{-\; \eta(t_i)\, g}$
\EndFor
\State \Return $z_{t_N}$
\end{algorithmic}
\end{algorithm}
\vspace{-2mm}
\cref{alg:omr} summarizes one full \methodname{} sampling pass. Relative to a standard flow-matching ODE solver, \methodname{} adds only an adapter forward pass, predictor-surprise evaluation, gradient computation, and final subtraction.

\vspace{-4mm}
\section{Experiments}
\vspace{-2mm}
\label{sec:exp}

\subsection{Setup}
\label{sec:exp-setup}

\paragraph{Generators and operating points.}
We use Wan2.2-T2V-A14B~\citep{wan22} as the primary text-to-video generator and refer to it as Wan2.2 below. Unless otherwise stated, Wan2.2 runs use $480 \times 832$ resolution, $81$ frames at $16$ FPS, $40$ ODE sampling steps, and CFG scales $4.0 / 3.0$ for the two experts. To test transfer, we also evaluate CogVideoX-5B at $480 \times 720$ and Wan2.2-I2V-A14B for image-to-video generation.

\vspace{-2mm}
\paragraph{World model.}
We use \vjepa{}~\citep{vjepa21} as the frozen world model. Both encoder and predictor heads are kept frozen throughout.

\vspace{-2mm}
\paragraph{Adapter.}
Wan2.2 uses the frozen $33.3$M adapter in \cref{sec:adapter}; CogVideoX uses the same architecture retrained on CogVideoX VAE-latent/\vjepa{}~\citep{vjepa21} pairs. Both are frozen during generation; \cref{tab:adapter-fidelity} evaluates the primary Wan2.2 adapter.

\vspace{-2mm}
\paragraph{Benchmarks.}
Our default evaluation uses our fixed $400$-prompt VideoPhy-2~\citep{videophy2} detailed subset, one generation per prompt. We fixed these $400$ prompt IDs from the released detailed-prompt pool before any generation or scoring, then used the identical list for every method with no output- or score-based filtering. We report SA, PC, and joint success (both scores $\geq4$ on the 1--5 scale). Wan2.2 efficiency and CogVideoX transfer use separate fixed $50$-prompt VideoPhy-2 subsets; Physics-IQ transfer uses a fixed $50$-prompt official I2V subset~\citep{motamed2026physicsiq} and evaluator.

\vspace{-2mm}
\paragraph{No-generator-fine-tuning comparisons.}
We compare Wan2.2~\citep{wan22} with three variants that leave the base generator frozen: CFG-Zero*~\citep{fan2025cfgzero}, P\&P~\citep{jang2026selfrefining}, and \methodname{}. CFG-Zero* modifies classifier-free guidance; P\&P uses generator self-consistency through predict-and-perturb refinement; \methodname{} instead uses adapter-space \vjepa{}~\citep{vjepa21} surprise inside the sampling trajectory. \Cref{tab:wallclock} reports the complementary wall-clock/search tradeoff, including best-of-$N$ and WMReward-style~\citep{yuan2026wmreward} search baselines.

\vspace{-2mm}
\paragraph{Best-of-$N$ selector.}
For the BoN rows in \cref{tab:wallclock}, we sample $N$ independent candidates from the corresponding base generator for the same prompt and select the candidate with the best scalar WMReward/\vjepa{}-style physical-plausibility proxy. The selector never sees VideoPhy-2 labels; this distinction matters because the proxy can prefer clips that look locally plausible while still failing the benchmark's physical-commonsense judgment.

\vspace{-2mm}
\subsection{Main results}
\label{sec:exp-headline}

\paragraph{Primary VideoPhy-2 comparison.}

\begin{table}[!t]
\centering
\caption{Frozen-generator test-time comparison on our fixed $400$-prompt VideoPhy-2~\citep{videophy2} detailed subset. Threshold columns report percentages over the same prompt set. We emphasize Physical-Commonsense (PC) because semantic adherence varies less across methods.}
\label{tab:videophy2}
\vspace{1mm}
\small
\setlength{\tabcolsep}{3.5pt}
\begin{tabular}{@{}lccccc@{}}
\toprule
Method & SA mean & SA$\geq4$ & \textbf{PC mean} & \textbf{PC$\geq4$} & Joint \\
\midrule
Wan2.2~\citep{wan22} & 3.740 & 68.0 & 3.660 & 57.0 & 47.00 \\
\quad + CFG-Zero*~\citep{fan2025cfgzero} & 3.770 & 68.0 & 3.610 & 56.0 & 48.25 \\
\quad + P\&P~\citep{jang2026selfrefining} & \textbf{3.780} & \textbf{70.0} & 3.650 & 55.0 & 48.75 \\
\midrule
\textbf{\methodname{} (ours)} & 3.750 & 68.0 & \textbf{3.700} & \textbf{60.0} & \textbf{52.00} \\
\bottomrule
\end{tabular}
\vspace{-2mm}
\end{table}

\Cref{tab:videophy2} shows that \methodname{} improves the joint metric from $47.00\%$ to $52.00\%$ ($+5.00$pp absolute; $+10.6\%$ relative) and exceeds the strongest listed frozen-generator baseline, P\&P, by $3.25$pp. The gain is physical rather than semantic: PC mean rises from $3.660$ to $3.700$ and PC$\geq4$ rises from $57.0\%$ to $60.0\%$, while SA$\geq4$ remains $68.0\%$. The per-category results in the supplementary material (\cref{app:category-breakdown}) show the largest joint improvements on contact dynamics and fluid/deformable motion.

\paragraph{Efficiency and cross-generator transfer.}

\begin{table}[!t]
\centering
\caption{Quality--efficiency on fixed $50$-prompt VideoPhy-2~\citep{videophy2} subsets. Compare scores only within panels; time is relative to each base and VRAM is peak usage. First-panel Wan2.2/\methodname{} scores are three-seed means; others are point estimates.}
\label{tab:wallclock}
\label{tab:cogvideox}
\vspace{1mm}
\scriptsize
\setlength{\tabcolsep}{5.5pt}
\begin{tabular}{@{}lccc@{}}
\toprule
Method & Relative time & Peak VRAM (GB) & Joint (\%, $\Delta$) $\uparrow$ \\
\midrule
\multicolumn{4}{@{}l}{\textit{Wan2.2-T2V-A14B, $480{\times}832$, fixed $50$-prompt efficiency subset}} \\
Wan2.2 & $1.00\times$ & 65.4 & 48.0 \\
\quad + P\&P & $1.6\times$ & 72.7 & 50.6 ($+2.6$) \\
\quad + BoN-2 & $2.00\times$ & 65.4 & 50.4 ($+2.4$) \\
\quad + WMReward-$\nabla$+BoN-4 & $19.88\times$ & 134 & 53.6 ($+5.6$) \\
\textbf{\quad + \methodname{} (ours)} & $1.71\times$ & 74.1 & \textbf{54.0 ($+6.0$)} \\
\midrule
\multicolumn{4}{@{}l}{\textit{CogVideoX-5B, $480{\times}720$, fixed $50$-prompt transfer subset}} \\
CogVideoX-5B & $1.00\times$ & 23.4 & 32.0 \\
\textbf{\quad + \methodname{} (ours)} & $1.67\times$ & 30.4 & \textbf{42.0 ($+10.0$)} \\
\quad + BoN-4 & $4.00\times$ & 23.4 & 36.0 ($+4.0$) \\
\quad + WMReward-$\nabla$ & $4.97\times$ & 83.2 & 38.0 ($+6.0$) \\
\quad + WMReward-$\nabla$+BoN-4 & $19.88\times$ & 83.2 & 40.0 ($+8.0$) \\
\quad + WMReward-$\nabla$+BoN-8 & $39.76\times$ & 83.2 & 40.0 ($+8.0$) \\
\bottomrule
\end{tabular}
\vspace{-2mm}
\end{table}

On the fixed $50$-prompt Wan2.2 efficiency subset, \methodname{} achieves the best Joint score, $54.0\%$ ($+6.0$pp), at $1.71\times$ runtime and $74.1$~GB peak VRAM.\par\noindent WMReward-$\nabla$ with BoN-4 reaches $53.6\%$ ($19.88\times$, 134~GB)---$11.6\times$ slower and $80.8\%$ more memory---while P\&P and BoN-2 trail by $3.4/3.6$pp. These subset results complement the $400$-prompt result in \cref{tab:videophy2}.

On CogVideoX, \methodname{} raises Joint from $32.0\%$ to $42.0\%$ ($1.67\times$ runtime, 30.4~GB peak VRAM). WMReward-$\nabla$ reaches $38.0\%$ ($4.97\times$, 83.2~GB); its BoN-4/8 variants reach $40.0\%$ at $19.88\times/39.76\times$ runtime, respectively.

WMReward-$\nabla$ differentiates through a V-JEPA encoder and VAE decoder~\citep{yuan2026wmreward}; \methodname{} uses only the predictor and generator-specific adapter, avoiding two additional Jacobians and their compute. \methodname{}'s larger matched CogVideoX gain ($+10.0$pp vs. $+6.0$pp) is consistent with a useful direct correction, but gradient-norm measurements would be needed to show that its gradients are larger or better preserved. \methodname{} remains complementary to candidate search.

\begin{figure}[H]
\centering
\vspace{-2mm}
\includegraphics[width=0.95\linewidth]{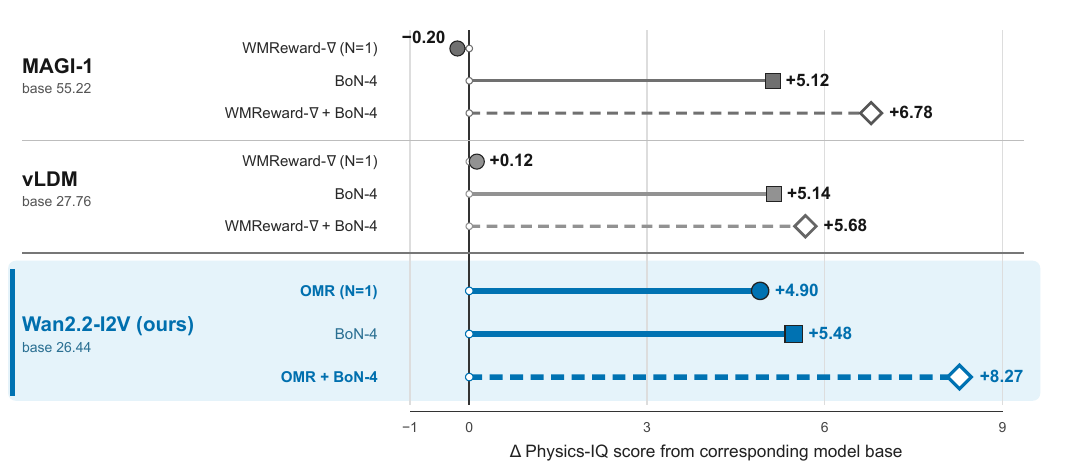}
\caption{\textbf{Physics-IQ gain from each base on one shared scale.} Gray gives published mixed-modality WMReward context: MAGI-1 is V2V and vLDM is I2V. Blue is our fixed-$50$-prompt Wan2.2-I2V pilot. Circles, squares, and diamonds denote guidance, BoN-4, and their combination; exact scores are supplementary (\cref{tab:physics-iq-values}).}
\label{fig:physics-iq}
\vspace{-2mm}
\end{figure}

\vspace{-2mm}
\paragraph{Cross-benchmark transfer on Physics-IQ.}
We further extend \methodname{} to image-to-video generation using the Wan2.2-I2V-A14B checkpoint and evaluate it on the Physics-IQ benchmark. \Cref{fig:physics-iq} compares our Wan2.2-I2V configurations with the published MAGI-1 (V2V) and vLDM (I2V) WMReward configurations~\citep{yuan2026wmreward}.

On Wan2.2-I2V, \methodname{} improves Physics-IQ by $+4.90$ without an in-loop VAE decode or \vjepa{} encoder pass and reaches $+8.27$ with BoN-4. Published single-trajectory WMReward changes are $-0.20/+0.12$ for MAGI-1 (V2V)/vLDM (I2V). The controlled Wan2.2 comparison supports one-trajectory correction complementary to selection; the gray mixed-modality references remain unmatched context. We treat it as transfer evidence; \cref{sec:adapter-fidelity} evaluates alignment.
\begin{figure}[p]
\centering
\includegraphics[height=0.86\textheight,keepaspectratio]{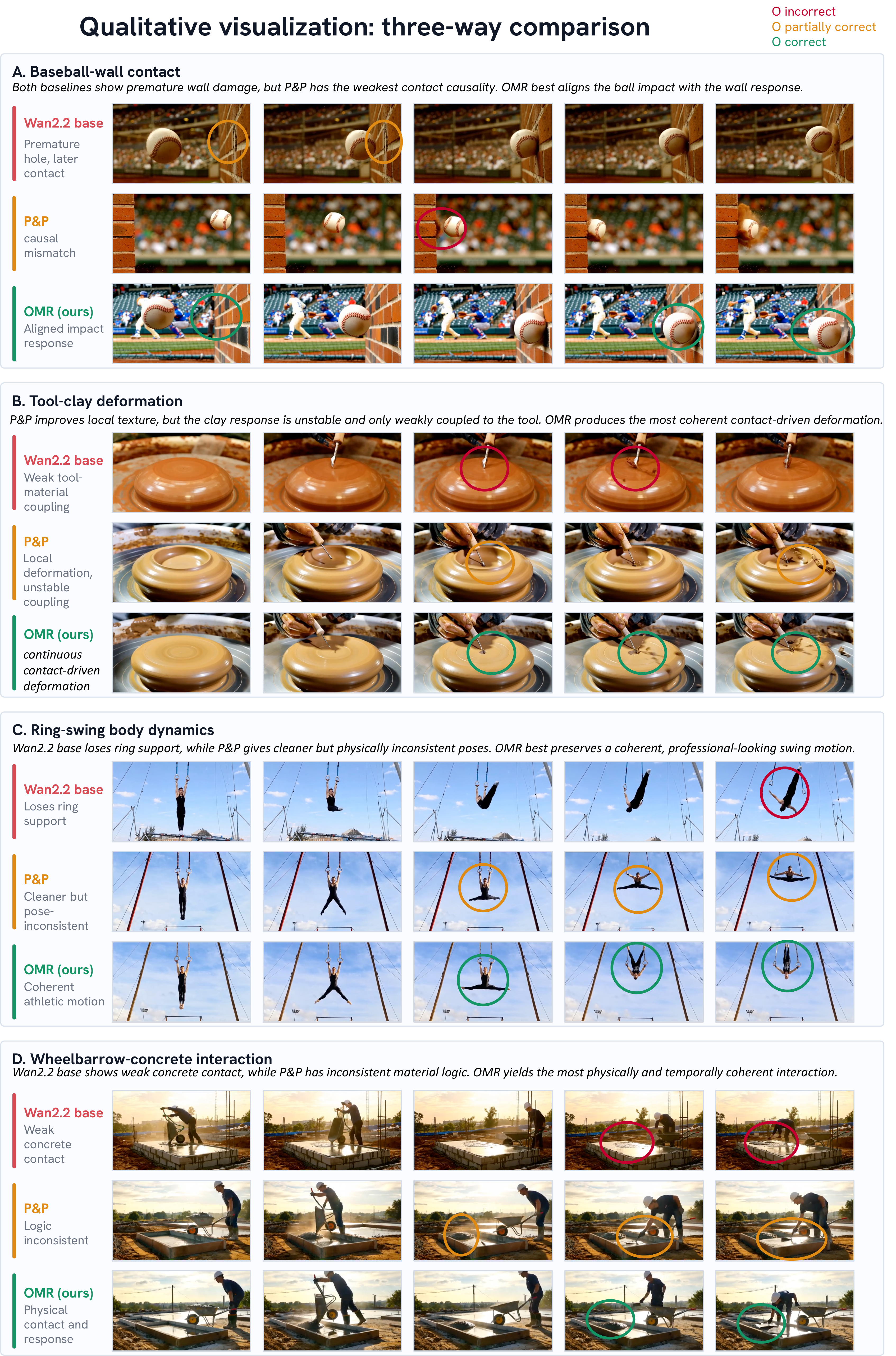}
\caption{\textbf{Qualitative visualization.} Four frame-strip examples compare Wan2.2~\citep{wan22}, P\&P~\citep{jang2026selfrefining}, and \methodname{} (ours). The examples emphasize temporal cause and effect: impact should coincide with material response, tools should stay coupled to the object they deform, and body motion should preserve support and momentum across frames.}
\label{fig:qualitative-main}
\end{figure}
\FloatBarrier

\subsection{Qualitative and human evaluation}
\label{sec:qualitative-main}

\Cref{fig:qualitative-main} diagnoses the failures behind the aggregate scores. In \textbf{A}, Wan2.2 shows a hole before clear impact and P\&P shows damage without synchronized contact; \methodname{} best aligns impact and response. In \textbf{B}, Wan2.2 weakly couples tool and clay, while P\&P's deformation is intermittent; \methodname{} maintains contact and accumulates deformation. In \textbf{C}, Wan2.2 loses ring support and P\&P produces an implausible leg transition; \methodname{} gives the most support- and momentum-consistent swing. In \textbf{D}, \methodname{} better preserves tool--concrete contact and coherent material response.

\paragraph{Human preference against P\&P.}
\label{sec:human-preference}

We compare \methodname{} against P\&P~\citep{jang2026selfrefining} in a blind study on $100$ detailed VideoPhy-2~\citep{videophy2} prompts. Thirteen annotators provide $1{,}300$ judgments per question; matching, randomization, and raw counts are in the supplementary material (\cref{app:human-study}).

\definecolor{omrwin}{RGB}{37,99,235}
\definecolor{prefTie}{RGB}{150,150,150}
\definecolor{srwin}{RGB}{234,88,12}

\newcommand{\prefrowmain}[5]{%
  \pgfmathsetmacro{\tieStart}{#3}
  \pgfmathsetmacro{\srStart}{#3 + #4}
  \node[anchor=east,align=right,font=\scriptsize] at (-2,#1+0.18) {#2};
  \draw[fill=omrwin!22,draw=white] (0,#1) rectangle (#3,#1+0.36);
  \draw[fill=prefTie!30,draw=white] (\tieStart,#1) rectangle (\srStart,#1+0.36);
  \draw[fill=srwin!22,draw=white] (\srStart,#1) rectangle (100,#1+0.36);
  \draw[black!35] (0,#1) rectangle (100,#1+0.36);
  \node[font=\scriptsize,text=omrwin!85!black] at (#3/2,#1+0.18) {#3\%};
  \node[font=\scriptsize,text=black!65] at ({#3 + #4/2},#1+0.18) {#4\%};
  \node[font=\scriptsize,text=srwin!85!black] at ({#3 + #4 + #5/2},#1+0.18) {#5\%};
}

\begin{figure}[H]
\centering
\begin{tikzpicture}[x=0.0068\linewidth,y=0.58cm]
  \prefrowmain{2.4}{Physical plausibility}{49.69}{30.08}{20.23}
  \prefrowmain{1.5}{Prompt adherence}{25.08}{45.00}{29.92}
  \prefrowmain{0.6}{Pooled judgments}{37.38}{37.54}{25.08}

  \foreach \x/\lab in {0/0,25/25,50/50,75/75,100/100} {
    \draw[black!25] (\x,0.42) -- (\x,2.94);
    \node[font=\scriptsize,text=black!60] at (\x,0.18) {\lab};
  }
  \node[font=\scriptsize,text=black!70] at (50,-0.18) {Pooled judgment share (\%)};

  \draw[fill=omrwin!22,draw=white] (13,3.28) rectangle (18,3.52);
  \node[anchor=west,font=\scriptsize] at (19,3.40) {\methodname{} wins};
  \draw[fill=prefTie!30,draw=white] (42,3.28) rectangle (47,3.52);
  \node[anchor=west,font=\scriptsize] at (48,3.40) {Tie};
  \draw[fill=srwin!22,draw=white] (61,3.28) rectangle (66,3.52);
  \node[anchor=west,font=\scriptsize] at (67,3.40) {P\&P wins};
\end{tikzpicture}
\vspace{-4mm}
\caption{\textbf{Human preference against P\&P~\citep{jang2026selfrefining}.} Pooled shares on $100$ matched prompts from $13$ annotators. \methodname{} wins physical plausibility ($49.69\%$ vs. $20.23\%$); ties are the largest prompt-adherence outcome. Protocol and counts are in the supplementary material (\cref{app:human-study}).}
\label{fig:human-preference}
\vspace{-4mm}
\end{figure}
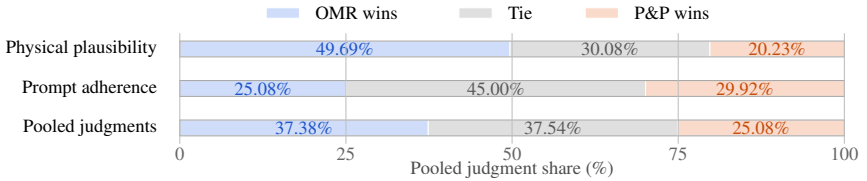

\Cref{fig:human-preference} shows a clear physical-plausibility advantage. Ties are the largest prompt-adherence outcome, with a small P\&P edge among non-tied judgments; the descriptive pool across both questions favors \methodname{}.

\vspace{-2mm}
\subsection{Discussion}
\label{sec:adapter-fidelity}

We ask whether a frozen world model can control a separately trained generator without joint training, and whether an adapter trained on real-video latents remains reliable during sampling.

\paragraph{Why can independently trained representations be mapped?}
The Wan latent and \vjepa{} target encode the same video under different objectives: one retains content and motion for synthesis, the other structure for future prediction. A task-specific transformation is therefore plausible without requiring identical or globally aligned spaces. The adapter learns this mapping into \vjepa{} token space while preserving the geometry and surprise ordering used by the predictor. This agrees with the intuition of the Platonic Representation Hypothesis~\citep{huh2024platonic}, but our evidence supports only this operational correspondence, not universal equivalence.

\paragraph{Can a real-video-trained adapter remain reliable during noisy sampling?}
The adapter is trained on real OpenVid-1M~\citep{nan2025openvid} Wan latents but receives provisional generated clean predictions $\zhone$, not raw noisy states $\zt$. We compare it with the exact decode--encode path on $500$ held-out real clips, four scheduled predictions from each of $50$ trajectories ($200$ states), and one matched finished control per trajectory ($50$ controls). This pairing isolates the extra mid-trajectory shift from the ordinary real-to-generated shift.

\begin{table}[H]
\centering
\caption{Adapter fidelity against the exact VAE decode-and-encode path using \vjepa{}~\citep{vjepa21}. Token cosine measures representation agreement; rank metrics compare surprise ordering. Higher is better.}
\label{tab:adapter-fidelity}
\vspace{1mm}
\small
\setlength{\tabcolsep}{4pt}
\begin{tabular}{@{}lcccc@{}}
\toprule
Evaluation point & $N$ & Token cosine & Spearman $\rho$ & Kendall $\tau$ \\
\midrule
Held-out real-video clips & $500$ & 0.7506 & 0.668 & 0.477 \\
Mid-trajectory clean predictions & $200$ & 0.762 & 0.617 & 0.446 \\
Matched finished controls & $50$ & 0.769 & 0.624 & 0.451 \\
\bottomrule
\end{tabular}
\vspace{-1mm}
\end{table}

Mid-trajectory cosine/$\rho$/$\tau$ ($0.762/0.617/0.446$) remains close to matched finished controls ($0.769/0.624/0.451$), so the adapter preserves an actionable representation and surprise ordering where \methodname{} operates. This does not establish robustness to raw-noise states, exact adapter--encoder equivalence, elimination of the real-to-generated shift, or causal attribution of the downstream gain. Protocol and slice diagnostics are supplementary (\cref{app:midtrajectory-fidelity}).

\vspace{-3mm}

\section{Conclusion}
\label{sec:conclusion}
\label{sec:limitations}

\noindent\textbf{Contribution and findings.}
\methodname{} shows that a frozen video world model can guide a separately trained generator within one sampling trajectory. A generator-specific adapter maps provisional clean latents into \vjepa{} feature space, using masked-future surprise as a dense correction before decoding. The generator and world model remain frozen, with no in-loop VAE decoding or video encoding. \methodname{} improves Wan2.2 from $47.0\%$ to $52.0\%$ on our fixed $400$-prompt VideoPhy-2 subset, at $1.71\times$ runtime in a separate $50$-prompt study. CogVideoX, Physics-IQ, and mid-trajectory fidelity results support transfer beyond the primary generator and modality.

\noindent\textbf{Limitations.}
\methodname{} inherits the blind spots and data biases of its world model; predictive plausibility does not guarantee physical correctness. Evaluation covers one world-model family, two generators, and the resolution used to train each adapter, while a new VAE latent space requires retraining. The $400$-prompt result uses one generation per prompt, and the $50$-prompt pilots are point estimates without paired prompt-level confidence intervals. Fidelity establishes representation and ranking agreement, but not exact adapter--encoder equivalence, raw-noise robustness, or causal attribution. \methodname{} also adds runtime and memory relative to base sampling.

\noindent\textbf{Future work.}
Promising directions include shared multi-resolution adapters, longer videos, complementary physical predictors, and adaptive uncertainty-driven schedules. Larger repeated-seed evaluations, gradient diagnostics, and causal interventions would better characterize failure modes. Combining \methodname{} with a small final selection budget may further improve physical consistency without returning to expensive large-pool search.

\section*{Acknowledgments}
We gratefully acknowledge support from the NAIRR Pilot, which provided compute resources under allocation NAIRR260204 for testing the core idea of this work.

\clearpage
\bibliography{references}

\clearpage
\appendix
\phantomsection
\label{app:supplementary}
\begin{center}
{\Large\bfseries Supplementary Material}\\[0.6em]
{\large\bfseries Off-Manifold Refinement: Guiding Video Generators with a Frozen World Model}\\[0.35em]
  {\normalsize Hai Nguyen-Truong, Tuan-Anh Vu, and Dang Huynh}
\end{center}
\vspace{1em}
\section{Adapter architecture details}
\label{app:adapter}

\subsection{Block diagram and tensor shapes}
\label{app:adapter-block}

The latent-to-embedding adapter $\Aps$ is a $33.3$M-parameter convolutional network, not a transformer. It trilinearly interpolates Wan2.2~\citep{wan22} VAE latents to the \vjepa{}~\citep{vjepa21} token grid, applies a local 3D convolutional stack, and then uses a shared per-token MLP head to predict hierarchical \vjepa{} embeddings. The trained Wan2.2 adapter uses $512$ convolution channels, an MLP hidden width of $2048$, three convolution layers, grid interpolation, and an output dimension of $6656$. The main paper describes the training data and objective; this appendix provides the architecture and parameter breakdown.

\begin{figure}[H]
\centering
\begin{tikzpicture}[
  box/.style={draw=black!35,rounded corners=1pt,align=center,inner sep=4pt,text width=0.88\linewidth,font=\small},
  arr/.style={->,thick,draw=black!55}
]
\node[box] (in) at (0,0) {$z \in \mathbb{R}^{B \times 16 \times 21 \times 60 \times 104}$\\Wan2.2 latent for $81{\times}480{\times}832$ video};
\node[box] (interp) at (0,-1.35) {Trilinear interpolation to the \vjepa{} grid\\$(21,60,104) \rightarrow (24,24,24)$\\output: $\mathbb{R}^{B \times 16 \times 24 \times 24 \times 24}$};
\node[box] (conv) at (0,-3.05) {3D convolutional stack\\Conv3d $16\!\rightarrow\!512$, Conv3d $512\!\rightarrow\!512$, Conv3d $512\!\rightarrow\!512$\\kernel $3^3$, padding $1$, SiLU after each conv\\output: $\mathbb{R}^{B \times 512 \times 24 \times 24 \times 24}$};
\node[box] (flat) at (0,-4.75) {Flatten space-time grid to tokens\\$\mathbb{R}^{B \times 13824 \times 512}$, where $13824 = 24\cdot24\cdot24$};
\node[box] (mlp) at (0,-6.25) {Shared per-token MLP head\\Linear $512\!\rightarrow\!2048$, Linear $2048\!\rightarrow\!2048$, Linear $2048\!\rightarrow\!6656$\\output: $\hat{y} \in \mathbb{R}^{B \times 13824 \times 6656}$};
\draw[arr] (in) -- (interp);
\draw[arr] (interp) -- (conv);
\draw[arr] (conv) -- (flat);
\draw[arr] (flat) -- (mlp);
\end{tikzpicture}
\caption{\textbf{Wan2.2 latent-to-\vjepa{} adapter.} The adapter combines convolutional layers with a shared per-token MLP head. It contains no self-attention layers; the \vjepa{} encoder already defines the target token grid.}
\label{fig:adapter-block}
\end{figure}
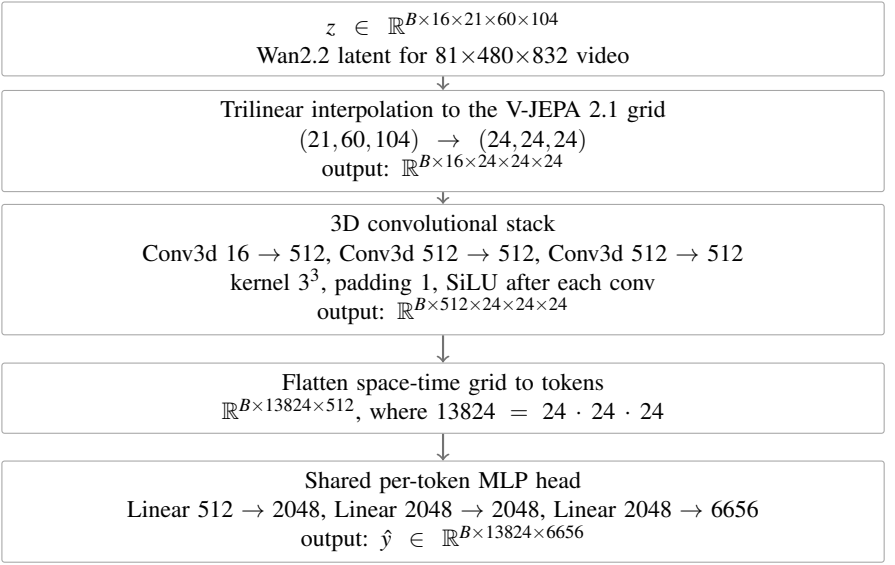

The adapter maps the $81$-frame Wan2.2 latent to the space--time grid of its precomputed \vjepa{} target. \vjepa{} evaluates a $48$-frame window from the generated video; within that window, visible context tokens condition predictions at masked future positions, and surprise is measured only on those masked targets. Dividing the $48$ input frames by tubelet size $2$ gives $24$ temporal tokens, while dividing image size $384$ by patch size $16$ gives a $24 \times 24$ spatial grid. The same windowing and masking rule is used when computing exact encoder targets and adapter-space surprise. The output dimension $D=6656$ is the hierarchical target used by the trained Wan2.2 adapter: four distillation layers, each of dimension $1664$. A single-layer target with $D=1664$ is supported by the code, but is not the variant used for the main experiments.

\paragraph{CogVideoX adapter.}
The cross-generator experiment does not reuse the Wan2.2 adapter weights. We retrain the same $33.3$M architecture on cached CogVideoX VAE latents paired with frozen \vjepa{} encoder targets from the same OpenVid-1M~\citep{nan2025openvid} clip pool, using the objective in the main paper, and freeze it before evaluation. This isolates transfer of the OMR procedure rather than assuming that two generators share a VAE latent coordinate system.

\subsection{Parameter count}
\label{app:adapter-params}

\begin{table}[H]
\centering
\caption{Parameter count for the trained Wan2.2 adapter $\Aps$.}
\label{tab:adapter-params}
\small
\setlength{\tabcolsep}{4pt}
\begin{tabular}{llr}
\toprule
Block & Layer / shape & Params \\
\midrule
Conv stack & Conv3d $16\rightarrow512$, $3^3$ + bias & 221,696 \\
 & Conv3d $512\rightarrow512$, $3^3$ + bias & 7,078,400 \\
 & Conv3d $512\rightarrow512$, $3^3$ + bias & 7,078,400 \\
\cmidrule(lr){2-3}
 & Subtotal & 14,378,496 \\
\midrule
MLP head & Linear $512\rightarrow2048$ + bias & 1,050,624 \\
 & Linear $2048\rightarrow2048$ + bias & 4,196,352 \\
 & Linear $2048\rightarrow6656$ + bias & 13,638,144 \\
\cmidrule(lr){2-3}
 & Subtotal & 18,885,120 \\
\midrule
Total &  & 33,263,616 \\
\bottomrule
\end{tabular}
\end{table}

The MLP head accounts for $18.89$M parameters, or roughly $57\%$ of the adapter. This is mostly a consequence of the hierarchical output dimension $D=6656$, not a deliberate increase in model capacity. If the last projection predicts a single \vjepa{} layer instead ($D=1664$), the total parameter count drops to roughly $23.0$M. The convolutional architecture exploits the target's regular token grid and remains lightweight enough for evaluation inside the sampling loop. The reported fidelity establishes that this configuration preserves an actionable signal; it does not establish that convolution is optimal or that attention would provide no benefit.

\paragraph{Reproducing the count.}
Enumerating trainable parameters for the configuration above gives the total in \cref{tab:adapter-params}. A forward-shape check with $z\in\mathbb{R}^{1\times16\times21\times60\times104}$ returns $\hat{y}\in\mathbb{R}^{1\times13824\times6656}$.

\subsection{Adapter objective details}
\label{app:adapter-loss}

The relational term in the main paper's adapter loss is not an additional physics reward and does not use labels. It is a representation-distillation regularizer. The matrix $\mathrm{Sim}(e)$ records the internal geometry of the frozen \vjepa{} representation: which space-time patches are close to one another, which are far apart, and how context and future tokens are arranged relative to each other. Matching $\mathrm{Sim}(\hat e)$ to $\mathrm{Sim}(e)$ therefore asks the adapter to preserve the pairwise neighborhood structure among tokens, not merely fit each token independently.

This matters because \methodname{} feeds the adapter output into the frozen \vjepa{} predictor. If the true encoder places a moving-object token close to its future continuation and far from static background, the adapter should preserve that relation; otherwise the predictor may see an artificial context--future mismatch and produce a misleading surprise gradient. The coefficient on the relational term is kept small so that it stabilizes geometry while the dominant supervision remains direct embedding reconstruction and token-wise cosine alignment.

\section{Mid-trajectory adapter-fidelity protocol}
\label{app:midtrajectory-fidelity}

\paragraph{Held-out real-video clips.}
The separate held-out evaluation in the main paper uses $500$ real-video clips at $81$ frames and $480{\times}832$. The adapter reaches a mean per-token cosine of $0.7506$ against the exact decode-and-encode \vjepa{}~\citep{vjepa21} targets. The four hierarchical distillation slices obtain cosine scores of $0.750$, $0.807$, $0.837$, and $0.665$, so no slice collapses. Adapter-space and exact-path predictive surprise have Spearman $\rho=0.668$ and Kendall $\tau=0.477$. Positive rank correlations mean that clips assigned higher surprise by the exact path tend also to receive higher adapter-space surprise; values near zero would make the ordering uninformative, while negative values would invert the intended guidance signal.

\paragraph{Deployment-time protocol.}
The adapter is trained on VAE latents of real videos but is used on predicted clean latents during generation. To measure that deployment-time shift directly, we retain four guided mid-trajectory predicted clean latents $\zhone$ across the symbolic guidance band from each of $50$ sampling trajectories, giving $200$ state-level evaluations, plus one corresponding finished latent per trajectory, giving $50$ finished-latent evaluations. For every retained latent $z$, we construct the exact reference embedding by decoding $z$ through the frozen generator VAE and encoding the resulting pixels with the frozen \vjepa{} encoder. Token cosine measures direct agreement between this target and $\Aps(z)$. Spearman $\rho$ and Kendall $\tau$ measure whether predictive-surprise rankings from the adapter path agree with rankings from the exact decode-and-encode path within each evaluation set.

The mid-trajectory predicted clean latents obtain cosine/$\rho$/$\tau$ of $0.762/0.617/0.446$, compared with $0.769/0.624/0.451$ for the corresponding finished latents. The proximity of the two rows supports the adapter's use on OMR's operating distribution. This is a stability diagnostic, not evidence that the real-to-generated distribution shift disappears or that the adapter exactly reproduces the encoder.
\section{Per-category VideoPhy-2 breakdown}
\label{app:category-breakdown}

\begin{table}[H]
\centering
\caption{Per-category VideoPhy-2~\citep{videophy2} detailed-prompt breakdown for our fixed $400$-prompt subset. Entries are percentages computed from integer counts over the category size.}
\label{tab:category-breakdown}
\scriptsize
\setlength{\tabcolsep}{2.5pt}
\begin{tabular}{@{}lrrrrrrr@{}}
\toprule
Category & \# prompts & \multicolumn{3}{c}{Wan2.2} & \multicolumn{3}{c}{\methodname{}} \\
\cmidrule(lr){3-5}\cmidrule(l){6-8}
 & & SA$\geq4$ & PC$\geq4$ & Joint & SA$\geq4$ & PC$\geq4$ & Joint \\
\midrule
Contact dynamics & 128 & 68.0 & 57.0 & 44.5 & \textbf{70.3} & \textbf{60.2} & \textbf{55.5} \\
Fluid / deformable motion & 41 & 65.9 & 56.1 & 43.9 & \textbf{68.3} & \textbf{63.4} & \textbf{53.7} \\
Gravity / free fall & 40 & \textbf{70.0} & 60.0 & 60.0 & 65.0 & \textbf{62.5} & 60.0 \\
Object permanence & 2 & 50.0 & 50.0 & 50.0 & 50.0 & 50.0 & 50.0 \\
Other physics principles & 189 & \textbf{68.3} & 56.6 & 46.6 & 67.2 & \textbf{58.7} & \textbf{47.6} \\
\midrule
Total & 400 & 68.0 & 57.0 & 47.0 & 68.0 & \textbf{60.0} & \textbf{52.0} \\
\bottomrule
\end{tabular}
\end{table}

\Cref{tab:category-breakdown} shows broad physical-consistency gains, with the strongest joint-score improvements in local interaction categories. Contact dynamics improves from $44.5\%$ to $55.5\%$ joint score ($+11.0$pp), and fluid/deformable motion improves from $43.9\%$ to $53.7\%$ ($+9.8$pp). On PC$\geq4$, the largest gain is in fluid/deformable motion, where \methodname{} rises from $56.1\%$ to $63.4\%$ ($+7.3$pp) over Wan2.2.

The trend is not limited to those categories: the mixed-phenomena ``Other physics principles'' row improves from $46.6\%$ to $47.6\%$ joint score, with PC$\geq4$ rising from $56.6\%$ to $58.7\%$. Because this row mixes several physical phenomena, we use it to check that the aggregate does not collapse outside contact/deformation prompts rather than to isolate one mechanism. Gravity/free-fall also improves on PC$\geq4$ ($60.0\%$ to $62.5\%$) while tying on joint score, and the two-prompt object-permanence slice is unchanged and too small to interpret.

\section{Cross-benchmark and cross-generator pilot protocols}
\label{app:cross-pilot-protocols}

\paragraph{Physics-IQ.}
The Physics-IQ~\citep{motamed2026physicsiq} experiment uses a fixed $50$-prompt subset of the official image-to-video benchmark. Base, OMR, Wan2.2+BoN-4, and OMR+BoN-4 share the Wan2.2-I2V-A14B checkpoint, conditioning images, prompt subset, and official evaluator. The exact prompt list will be included with the release. The published WMReward~\citep{yuan2026wmreward} references use different modalities: MAGI-1 is V2V, whereas vLDM is I2V. They provide unmatched mixed-modality context only and are not OMR evaluations. Because our pilot covers only $50$ prompts, we report point estimates and do not present it as a full-benchmark result.

\begin{table}[H]
\centering
\caption{Exact values for the Physics-IQ comparison in the main paper. Published rows provide mixed-modality context; our matched pilot is I2V. $N$ is the number of generated candidates and $\Delta$ is relative to the corresponding base. Published WMReward-$\nabla$ repeatedly uses the VAE decode-and-encode path; \methodname{} uses the adapter and frozen predictor without an in-loop decode or encoder pass.}
\label{tab:physics-iq-values}
\small
\setlength{\tabcolsep}{4.5pt}
\begin{tabular}{@{}lclcc@{}}
\toprule
Model and method & $N$ & World-model use & Score $\uparrow$ & $\Delta$ \\
\midrule
\multicolumn{5}{@{}l}{\textit{Published WMReward mixed-modality context}} \\
MAGI-1 (V2V) & 1 & -- & 55.22 & -- \\
\quad + WMReward-$\nabla$ & 1 & Decode $\rightarrow$ encoder ($\nabla$) & 55.02 & $-0.20$ \\
\quad + BoN-4 & 4 & Completed-video selection & 60.34 & $+5.12$ \\
\quad + WMReward-$\nabla$ + BoN-4 & 4 & Decode $\rightarrow$ encoder ($\nabla$) + selection & \textbf{62.00} & $\mathbf{+6.78}$ \\
\cmidrule(lr){1-5}
vLDM (I2V) & 1 & -- & 27.76 & -- \\
\quad + WMReward-$\nabla$ & 1 & Decode $\rightarrow$ encoder ($\nabla$) & 27.88 & $+0.12$ \\
\quad + BoN-4 & 4 & Completed-video selection & 32.90 & $+5.14$ \\
\quad + WMReward-$\nabla$ + BoN-4 & 4 & Decode $\rightarrow$ encoder ($\nabla$) + selection & \textbf{33.44} & $\mathbf{+5.68}$ \\
\midrule
\multicolumn{5}{@{}l}{\textit{Our matched Wan2.2-I2V pilot (fixed $50$ prompts)}} \\
Wan2.2-I2V & 1 & -- & 26.44 & -- \\
\textbf{\quad + \methodname{} (ours)} & 1 & Adapter $\rightarrow$ predictor ($\nabla$) & 31.34 & $+4.90$ \\
\quad + BoN-4 & 4 & Completed-video selection & 31.92 & $+5.48$ \\
\textbf{\quad + \methodname{} + BoN-4 (ours)} & 4 & Adapter $\rightarrow$ predictor ($\nabla$) + selection & \textbf{34.71} & $\mathbf{+8.27}$ \\
\bottomrule
\end{tabular}
\end{table}

\paragraph{CogVideoX-5B.}
The cross-generator pilot uses CogVideoX-5B~\citep{yang2024cogvideox} at $480{\times}720$ on a fixed $50$-prompt VideoPhy-2~\citep{videophy2} subset. Its separately trained CogVideoX adapter is frozen for all OMR runs. Base, OMR, BoN-4, WMReward-$\nabla$, WMReward-$\nabla$+BoN-4, and WMReward-$\nabla$+BoN-8 use the same prompts, checkpoint, resolution, VideoPhy-2 evaluator, and hardware. Relative time is measured against the CogVideoX base row, and VRAM is peak allocated memory. These controls isolate the guidance/search path within the table, but the finite prompt subset remains subject to sampling variance.

\paragraph{Interpreting BoN.}
Physics-IQ scores spatiotemporal overlap with reference motion, so sample-and-select can benefit from choosing among multiple completed trajectories. Conversely, the VideoPhy-2 selector optimizes a world-model proxy rather than the final semantic-and-physical joint metric. This explains why a larger BoN pool need not monotonically improve the reported joint score. The complete prompt lists and all generated clips will be released so that both pilot evaluations can be audited.
\section{Relation to alternative inference-time refinement}
\label{app:method-comparison}

\methodname{} is deliberately simpler than predict-and-perturb refinement. It has no inner stochastic loop, mask blending, re-noising, or explicit re-projection step; sampling advances once from $t=0$ to $t=1$ under a modified velocity field. P\&P-style methods infer uncertainty by repeatedly querying the generator and measuring cross-sample disagreement, whereas \methodname{} obtains a first-order correction direction from one adapter-space world-model loss evaluation at each guided step. \methodname{} also differs from candidate-selection methods, which score completed clips, and WMReward-style gradient steering, which repeatedly decodes and re-encodes an intermediate clean estimate. Here, the world-model signal is applied before the sample is finished through the shorter adapter-space path.

\section{Seed stability check}
\label{app:seed-stability}

\begin{table}[H]
\centering
\caption{Seed-stability check on the fixed $50$-prompt Wan2.2 efficiency subset used in Table 2. Entries are mean $\pm$ standard deviation over three random seeds, with the same prompt list for every seed. This subset does not define the $400$-prompt aggregate in Table 1 of the main paper.}
\label{tab:seed-stability}
\small
\setlength{\tabcolsep}{5pt}
\begin{tabular}{@{}lccc@{}}
\toprule
Method & SA$\geq4$ & PC$\geq4$ & Joint \% \\
\midrule
Wan2.2~\citep{wan22} & $68.0 \pm 2.0$ & $56.7 \pm 1.2$ & $48.0 \pm 2.0$ \\
\textbf{\methodname{} (ours)} & $68.0 \pm 2.0$ & $\textbf{60.7} \pm 1.2$ & $\textbf{54.0} \pm 2.0$ \\
\bottomrule
\end{tabular}
\end{table}

\Cref{tab:seed-stability} checks whether the conclusion is dominated by seed choice on a smaller repeated-seed subset. A full $400$-prompt $\times$ $3$-seed evaluation would require $1{,}200$ high-resolution video generations per method, so we use this fixed $50$-prompt subset as a practical seed-sensitivity check. \methodname{} matches Wan2.2 on SA, improves PC by $+4.0$pp, and improves joint score by $+6.0$pp, with the same seed-to-seed standard deviation as Wan2.2 on all three threshold metrics. We therefore report the one-generation-per-prompt aggregate in Table 1 of the main paper as the main point estimate, with \cref{tab:seed-stability} serving as a sanity check rather than a formal confidence-interval study.

\section{Human preference protocol and raw counts}
\label{app:human-study}

We compare \methodname{} against P\&P~\citep{jang2026selfrefining} in a blind pairwise study on $100$ detailed VideoPhy-2~\citep{videophy2} prompts. Each pair uses a matched prompt, seed, resolution, frame count, and sampler budget, and presentation order is randomized. Thirteen annotators judge every pair separately for physical plausibility and prompt adherence, yielding $1{,}300$ judgments per question and $2{,}600$ judgments overall.

\begin{table}[H]
\centering
\caption{Raw human-preference counts. Columns give judgments favoring \methodname{}, ties, and judgments favoring P\&P. The pooled row combines both questions and is reported descriptively.}
\label{tab:human-raw-counts}
\small
\setlength{\tabcolsep}{7pt}
\begin{tabular}{@{}lrrrr@{}}
\toprule
Question & \methodname{} & Tie & P\&P & Total \\
\midrule
Physical plausibility & 646 & 391 & 263 & 1,300 \\
Prompt adherence & 326 & 585 & 389 & 1,300 \\
\midrule
Pooled & 972 & 976 & 652 & 2,600 \\
\bottomrule
\end{tabular}
\end{table}

\end{document}